\documentclass[letterpaper]{article} 
\usepackage[arxiv]{aaai24}  
\usepackage{times}  
\usepackage{helvet}  
\usepackage{courier}  
\usepackage[hyphens]{url}  
\usepackage{graphicx} 
\usepackage{natbib}  
\usepackage{caption} 
\usepackage{algorithm}
\usepackage{algorithmic}
\usepackage{xcolor}

\usepackage{newfloat}
\usepackage{listings}
\DeclareCaptionStyle{ruled}{labelfont=normalfont,labelsep=colon,strut=off} 
\floatstyle{ruled}
\newfloat{listing}{tb}{lst}{}
\floatname{listing}{Listing}
\title{IIDM: Inter and Intra-domain Mixing for Semi-supervised Domain Adaptation in Semantic Segmentation}
\author {
    Weifu Fu,
    Qiang Nie,
    Jialin Li,
    Yuhuan Lin,
    Kai Wu,
    Jian Li,
    Yabiao Wang,
    Yong Liu,
    Chengjie Wang
}
\affiliations {
    YouTu Lab, Tencent\\
    \{ryanwfu, stephennie, jarenli, gleelin, lloydwu, swordli, caseywang, choasliu, jasoncjwang\}@tencent.com
}

\usepackage{multirow}
\usepackage{amssymb}

\usepackage{bibentry}

\begin{document}

\maketitle

\begin{abstract}
Despite recent advances in semantic segmentation, an inevitable challenge is the performance degradation caused by the domain shift in real applications. Current dominant approach to solve this problem is unsupervised domain adaptation (UDA). However, the absence of labeled target data in UDA is overly restrictive and limits performance. To overcome this limitation, a more practical scenario called semi-supervised domain adaptation (SSDA) has been proposed. Existing SSDA methods are derived from the UDA paradigm and primarily focus on leveraging the unlabeled target data and source data. In this paper, we highlight the significance of exploiting the intra-domain information between the labeled target data and unlabeled target data. Instead of solely using the scarce labeled target data for supervision, we propose a novel SSDA framework that incorporates both Inter and Intra Domain Mixing (IIDM), where inter-domain mixing mitigates the source-target domain gap and intra-domain mixing enriches the available target domain information, and the network can capture more domain-invariant features. We also explore different domain mixing strategies to better exploit the target domain information. Comprehensive experiments conducted on the GTA5→Cityscapes and SYNTHIA→Cityscapes benchmarks demonstrate the effectiveness of IIDM, surpassing previous methods by a large margin.
\end{abstract}

\begin{figure}[t]
\centering
\includegraphics[width=0.98\columnwidth]{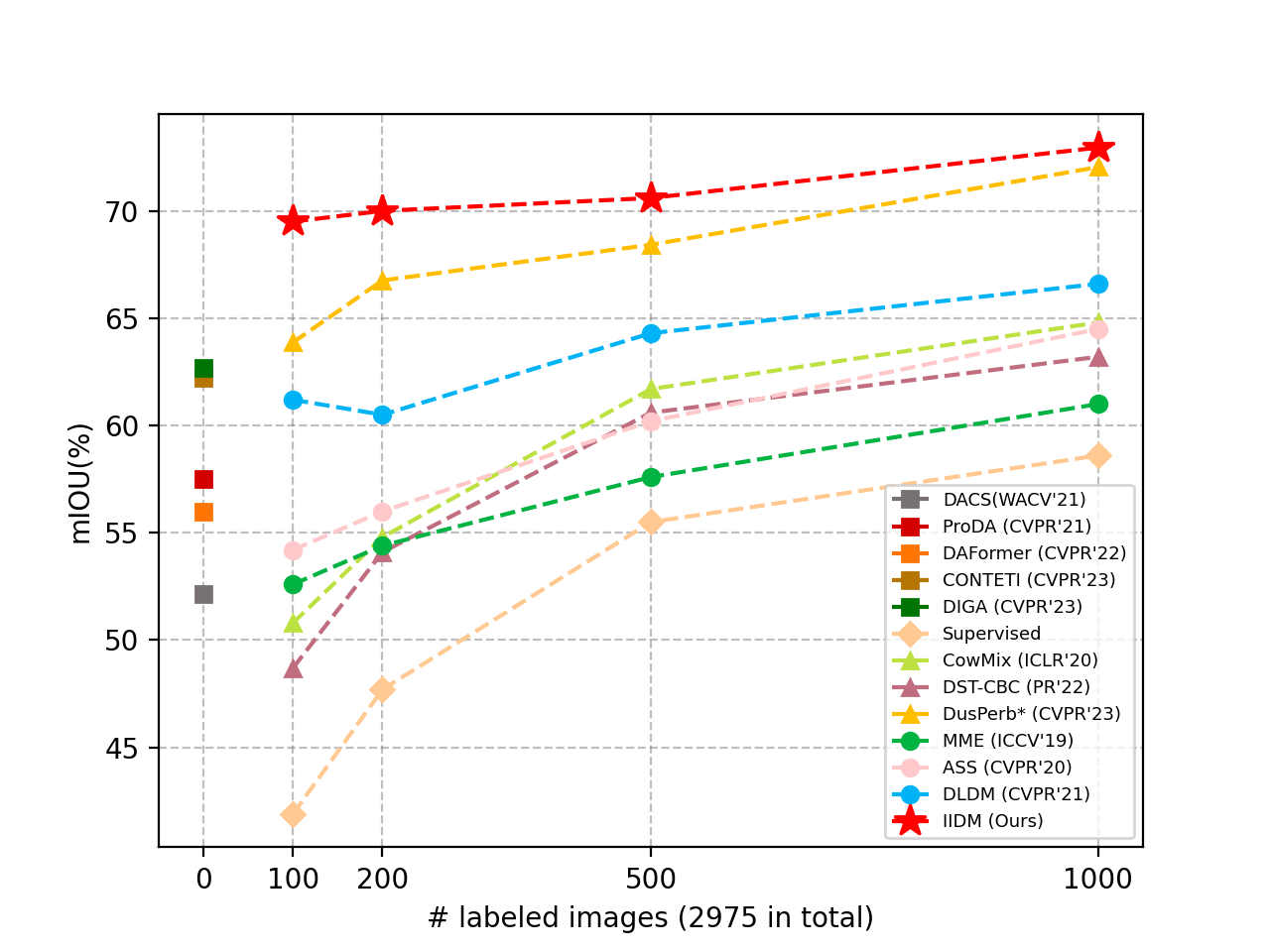} 
\caption{Comparison between state-of-the-art methods and the proposed Inter and Intra-Domain Mixing (IIDM) on the GTA5→Cityscapes setting. IIDM demonstrates superior performance across varying quantities of labeled target images, especially when the labeled target data is scarce.}
\label{comparison}
\end{figure}

\section{Introduction}
In recent years, there have been significant advancements in semantic segmentation models~\cite{chen2018encoder,xie2021segformer,zhou2022rethinking,cheng2022masked}. However, when deploying these models in real-world scenarios, performance degradation often occurs due to domain shift in the target data. To address this challenge, domain adaptation techniques have gained increasing interest in the field of deep learning-based semantic segmentation. These techniques aim to bridge the gap between training on source data and testing on target data, which typically exhibit domain differences.
Most current research focuses on the unsupervised domain adaptation (UDA) setting~\cite{zhang2021prototypical,hoyer2022daformer,shen2023diga,hoyer2023mic}, where only unlabeled target data is available alongside labeled source data. However, this assumption of having no labeled target data can be overly restrictive and may limit the performance of domain adaptation models. In practical scenarios, it is often feasible to label a small amount of target data. Therefore, a more practical setting called semi-supervised domain adaptation (SSDA) has been proposed, which leverages source data, a small amount of labeled target data, and a large amount of unlabeled target data. Compared to UDA, SSDA is important but seldom investigated in recent years.

Current researches on SSDA are influenced a lot by previous UDA methods, which emphasize more on leveraging the unlabeled target data and the source data. The small amount of labeled target data is always under exploited and merely trained as a supervision with cross-entropy (CE) loss. For example, \citeauthor{yu2023semi} proposed a label adaptation model to adapt the source label to the view of unlabeled target data based on pseudo labels. \citeauthor{chen2021semi2} proposed a cross-domain data mixing to mitigate the domain gap and a intra-domain data mixing solely involving unlabeled target data. \citeauthor{chen2021semi1} utilized a dual-level mixing between source data and unlabeled target data in region level and sample level. \citeauthor{qin2021contradictory} proposed to minimize the conditional entropy to cluster the features of unlabeled target data. The labeled target data is ignored because of its tiny amount compared to the unlabeled target data. All these methods does not fully utilize the relationship between the unlabeled target data and the labeled target data, which contains more discriminative information of the target domain.

\begin{figure}[t]
\centering
\includegraphics[width=0.98\columnwidth]{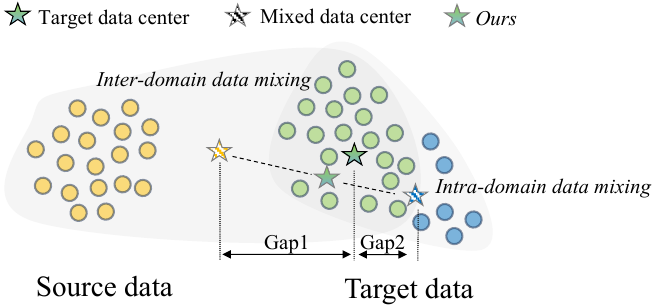}
\caption{Motivation for using both inter-domain and intra-domain data mixing, where Gap1 means the domain gap between inter-domain mixed data center and the real target data center. Gap2 denotes the gap between intra-domain mixed data center and the real target data center. Learning with both inter and intra-domain data mixing equals to the learning based on the pseudo data center (Ours) which is the closest to the real target data center.}
\label{motivation}
\end{figure}

As shown in Figure~\ref{motivation}, the large amount of unlabeled target data might well represents the target domain, where the center of the real target domain is represented by a green pentagram. 
However, without annotation, the unlabeled target data can not well leveraged for learning. To have a better performance on the target domain, previous domain adaptation methods utilize inter-domain data mixing between the source data and unlabeled target data, and obtain a mixed data center as shown by the orange pentagram, which is closer to the real target center. Thus, learning based on the mixed data with pseudo labels can lead to a performance promotion on target domain. However, learning based on the inter-domain mixed data still has a large gap with the learning on the real target data. Existing SSDA methods rely on the labeled target data as a supervision to improve the performance. As the amount of labeled target data is small, supervised learning on it contributes little on the final performance.


In this regard, we propose to simultaneously leveraging the Inter and Intra Domain Mixing (IIDM). As shown in Figure~\ref{motivation}, by mixing the labeled target data and unlabeled target data, an intra-domain mixed data center that is closer to the real domain center than inter-domain mixed center can be achieved. This is reasonable as semi-supervised learning (SSL) always has a better performance than UDA methods. With inter-domain data mixing and intra-domain data mixing, the learning can be deemed as the learning based on the pseudo data center denoted by the green pentagram (Ours) which is most near the real domain center. Attributed to inter-domain mixing and intra-domain mixing, the network can capture more domain-invariant features and promote its performance on the target domain. As shown in Figure~\ref{comparison}, IIDM achieves an significant promotion on SSDA performance compared to previous state-of-the-art (SOTA) methods, especially when the labeled target data is scarce. Our contribution can be summarized as follows:
\begin{itemize}
    \item We emphasize the importance of leveraging intra-domain data in a more effective manner within the context of semi-supervised domain adaptation (SSDA). Previous SSDA methods have often overlooked the potential benefits of utilizing intra-domain data.
    \item We propose a novel SSDA framework (IIDM) that emphasizes learning from both inter and intra-domain mixing. We also evaluate different domain mixing strategies to explore their impact on performance.
    \item The proposed IIDM achieves state-of-the-art performance on two widely used benchmarks, surpassing previous SSDA methods by a significant margin.
\end{itemize}

\section{Related Work}
\paragraph{Unsupervised domain adaptation (UDA) for semantic segmentation.} UDA mitigates the domain shift by exploiting the shared information between source and target domains. Two main branches of methodology are proposed to tackle the issue, namely adversarial training and pseudo-labelling. Adversarial training aims to align the distributions of the source and target domains at various levels, such as image~\cite{hoffman2018cycada}, feature~\cite{tsai2018learning,li2023contrast}, and prediction~\cite{vu2019advent}. In pseudo-labelling, the network learns by assigning pseudo-labels to unlabeled target data. The process of pseudo-labelling can be conducted either offline~\cite{zhang2021prototypical} or online~\cite{tranheden2021dacs,hoyer2022daformer}. To improve the learning of context relations, HRDA~\cite{hoyer2022hrda} uses multi-crop consistency training, and MIC~\cite{hoyer2023mic} use random patch masking to learn a larger set of different context clues. In our method, we adopt an online training structure, wherein the domain migration from the source to the target is accomplished by a simple and straightforward inter-domain mixing.

\paragraph{Semi-supervised learning (SSL) for semantic segmentation.} SSL methods in semantic segmentation can be grouped into consistency regularization and entropy minimization. Consistency regularization believes that applying different forms of perturbations to an image should have similar outputs. CowMix~\cite{french2019semi} figure out strong, high-dimensional perturbations are critical for consistency regularization in semi-supervised semantic segmentation, and similar strong perturbations include CutOut~\cite{devries2017improved} and CutMix~\cite{yun2019cutmix}. U2PL~\cite{wang2022semi} believes that uncertain pixels can be used as negative samples in contrast learning. UniMatch~\cite{yang2023revisiting} through dual-stream perturbation and feature-level perturbations to exploit a broader perturbation space. Entropy minimization is often employed in conjunction with self-training, where pseudo-labels are assigned to unlabeled data. ST++~\cite{yang2022st++} introduces a progressive retraining strategy based on image-level selection and highlights the effectiveness of strong perturbations in self-training. 
Our approach leverages the concept of consistent learning to effectively mitigate the intra-domain gap between labeled and unlabeled target data by employing intra-domain mixing.

\paragraph{Semi-supervised domain adaptation (SSDA) for semantic segmentation.} While significant progress has been made in the field of SSDA in image classification~\cite{saito2019semi,qin2021contradictory,yang2021deep,yu2023semi}, there is a noticeable lack of research focusing on applying SSDA techniques to semantic segmentation. ASS~\cite{wang2020alleviating} alleviates semantic-level shift through adversarial training to realize feature alignment across domain from global and semantic level. However, the adversarial loss used in this approach makes training unstable due to weak supervision, and the naive supervision of labeled target does not make full use of the source data. DLDM~\cite{chen2021semi1} incorporates a novel approach by utilizing two domain mixed teachers, corresponding to region-level and sample-level respectively.
Nevertheless, sample-level mixing and region-level mixing do not directly resolve the domain shift. Moreover, DLDM necessitates the involvement of two teachers and a two-stage training process, thereby introducing a heightened level of complexity. Intra-domain mixing between unlabeled target data are explored~\cite{chen2021semi2} to alleviate the domain shifting. 
However, the labeled and unlabeled target data are not thoroughly exploited. 
In our method, we addresses these limitations by introducing different inter and intra-domain mixing in an end-to-end framework.

\begin{figure*}[t]
\centering
\includegraphics[width=0.8\textwidth]{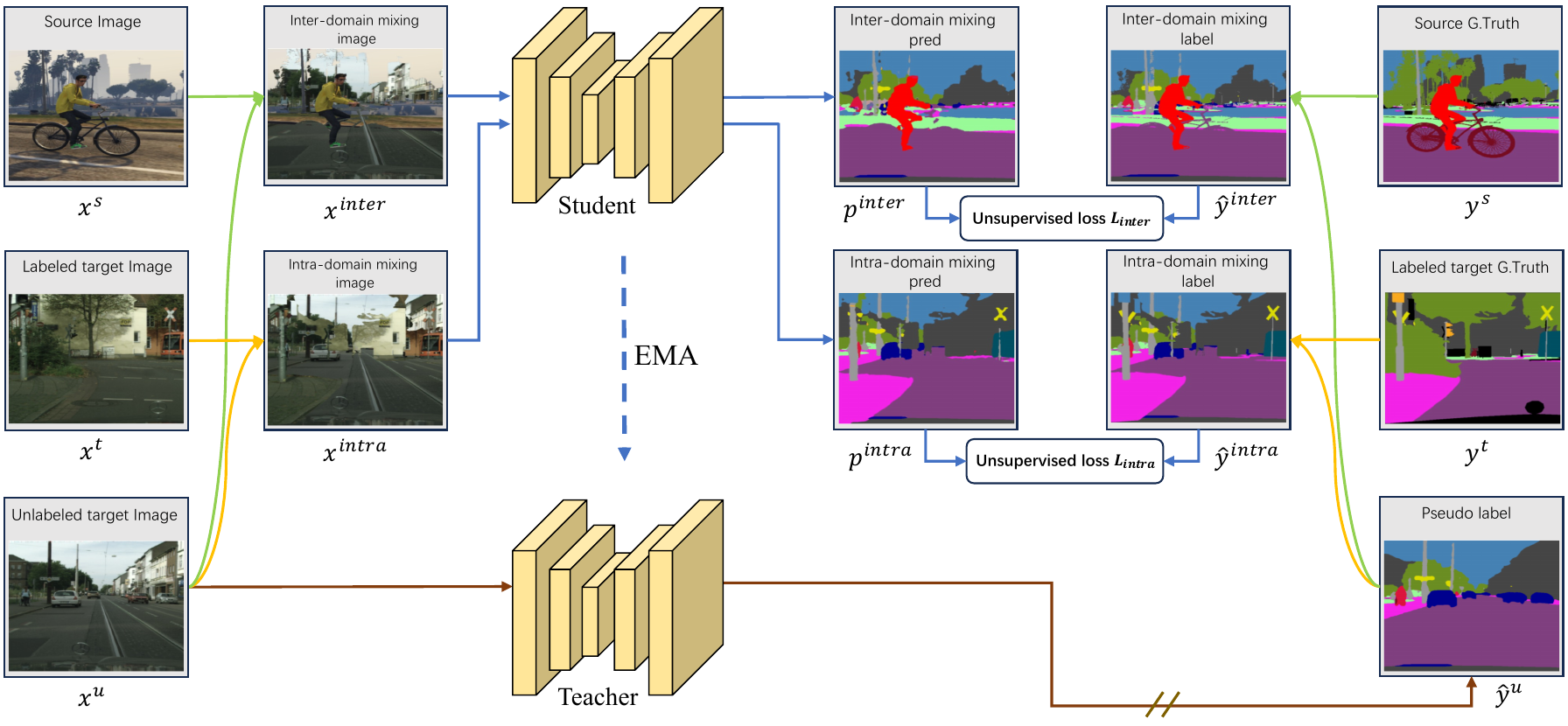} 
\caption{Overview of the proposed IIDM. Note that the supervised branches on source data and labeled target data are omitted for brevity. Inter-domain mixing (green line) is applied between source data and unlabeled target data, while intra-domain mixing (orange line) is applied between labeled target data and unlabeled target data. Additionally, pseudo-labels are generated through an exponential moving average (EMA) teacher model (brown line), based on the unlabeled target data without mixing.}
\label{overview}
\end{figure*}

\section{Method}
Previous research in SSDA has primarily focused on mitigating the domain shift between source and target domain, while labeled target data is mostly only used for supervised learning. In the proposed IIDM, we not only consider the conventional domain shift, the intra-domain gap between labeled target data and unlabeled target data due to random sampling is also taken into account, especially when the labeled target data is scarce, and a small set of that cannot represent the data of the entire target domain. Considering the above points, we mitigate the domain shift between source and target by employing inter-domain mixing. 
Additionally, we utilize intra-domain mixing that involves the mixing of labeled target and unlabeled target images to reduce the intra-domain gap.

\subsection{Preliminary}
In the setting of SSDA for semantic segmentation, a substantial volume of source domain data, a limited quantity of labeled target domain data, and a considerable amount of unlabeled target domain data are available.
Let $D_s = {(x_i^s, y_i^s)}_{i=1}^{N_s}$ represents the source data, where $x_i^s$ is source domain image, $y_i^s$ is the associated label of $x_i^s$, and $N_s$ is the number of labeled source images. Let $D_t = {(x_i^t, y_i^t)}_{i=1}^{N_t}$ and $D_u = {(x_i^u)}_{i=1}^{N_u}$ denote the labeled target data and unlabeled target data respectively, where $x_i^t$ is the labeled target image, $x_i^u$ is the unlabeled target image, $N_t$ and $N_u$ are the number of labeled and unlabeled target images correspondingly, and $y_i^t$ is the associated label of $x_i^t$. Leveraging these available data $(D_s, D_t, D_u)$, we hope to seek a model $g_{\theta}$ that exhibits good performance on the target domain.

\subsection{Inter-domain Mixing}
Inter-domain mixing is applied as a strong perturbation technique within the proposed framework on the source and target domain, as shown in Figure~\ref{overview}. 

In this paper, we choose ClassMix~\cite{olsson2021classmix} as the mixing technique.
The input of inter-domain mixing are source and unlabeled target data, and that is implemented as follow: (1) Source image $x_i^s$ and unlabeled target image $x_i^u$ are randomly sampled from $D_s$ and $D_u$ respectively. (2) $x_i^u$ with almost no disturbance is fed into the teacher network to get prediction $\hat{y}_i^u$, which can be regarded as pseudo label. (3) Randomly select half of the categories in $y_i^s$, and generate the binary mask $M_1$ corresponding to the half categories. The binary mask $M_1$ contains 1 values in chosen pixels and 0 values otherwise. (4) Utilize the binary mask $M_1$ with a value of 1 to denote the selection of pixels from $x_i^s$ and a value of 0 to indicate the selection of pixels from $x_i^u$. The selected pixels are used to generate inter-domain mixing image $x_i^{inter}$, and the generation process of $y_i^{inter}$ is similar. (5) $x_i^{inter}$ is fed into the student network to generate the mixed prediction.

The process of using $M_1$ to generate inter-domain mixing image can be formulated as:

\begin{equation}\label{eq:InterMix1}
    x_i^{inter}=M_1\odot x_i^s+(1-M_1)\odot x_i^u \\
\end{equation}
\begin{equation}\label{eq:InterMix2}
    \hat{y}_i^{inter}=M_1\odot y_i^s+(1-M_1)\odot \hat{y}_i^u
\end{equation}
where $\odot$ denotes element-wise multiplication. 

As shown in Figure~\ref{overview}, the inter-domain mixing image $x_i^{inter}$ contains the content of both $x_i^s$ and $x_i^u$. It work as a bridge to produce intermediate samples between different domains. Furthermore, this operation can destroy the inherent structure of the original image, thereby serving as a strong perturbation technique. 

\subsection{Intra-domain Mixing}
Intra-domain mixing also serves as a strong perturbation for consistent learning in the proposed method, and bridges the gap between the labeled and unlabeled target data.
The operational procedure bears a close resemblance to the process of inter-domain mixing. However, the distinguishing factor lies in the fact that the inputs originate exclusively from the same target domain.
The implementation details are as follow: (1) Labeled target image $x_i^t$ randomly sampled from $D_t$, and unlabeled target image $x_i^u$ is the same image in inter-domain mixing. (2) Get pseudo label $\hat{y}_i^u$ from inter-domain mixing procedure. (3) Randomly select half of the categories in $y_i^t$, and generate the binary mask $M_2$. (4) Utilize the binary mask $M_2$ to generate intra-domain mixing image $x_i^{intra}$ and corresponding label $y_i^{intra}$. (5) $x_i^{intra}$ is fed into the student network to generate the mixed prediction.

The process of using $M_2$ to generate intra-domain mixing image can be formulated as:

\begin{equation}\label{eq:IntraMix1}
    x_i^{intra}=M_2\odot x_i^t+(1-M_2)\odot x_i^u \\
\end{equation}
\begin{equation}\label{eq:IntraMix2}
    \hat{y}_i^{intra}=M_2\odot y_i^t+(1-M_2)\odot \hat{y}_i^u
\end{equation}

As shown in Figure~\ref{overview}, the intra-domain mixing image $x_i^{intra}$ contains the content of both $x_i^t$ and $x_i^u$. Due to random sampling, there is a domain gap between labeled target data and the whole target domain. We take this into consideration and reduce the gap in a similar manner of inter-domain mixing. Intra-domain mixing is equivalent to making another strong disturbance to $x_i^u$, further expanding the data distribution in the target domain.

\subsection{Holistic Framework}

On the basis of inter-domain mixing, we additionally propose intra-domain mixing, and better results can be achieved by combining those two domain mixing in SSDA. An overview of the proposed IIDM is shown in Figure~\ref{overview}.

We use the structure of mean teacher~\cite{tarvainen2017mean}. The teacher network $h_{\phi}$ has an identical architecture with the student network $g_{\theta}$, and the parameters are derived from the student network by exponential moving average (EMA) after each training step t.
\begin{equation}\label{eq:ema}
    \phi _{t+1} \gets \alpha \phi_{t} + (1-\alpha) \theta_{t+1} 
\end{equation}
The student network $g_{\theta}$ is trained on augmented images, while the teacher network $h_{\phi}$ generates the pseudo-labels using non-augmented images.


Teacher network is used for generating pseudo-labels and no gradients will be backpropagated. Additionally, a quality estimate is produced for the pseudo-labels. Here, we use the ratio of pixels exceeding a threshold $\tau$ of the maximum softmax probability~\cite{tranheden2021dacs} as quality $q_{i}$.

\begin{equation}\label{eq:quality}
    q_{i} = \frac{{{\textstyle \sum_{H\times W}^{j=1}} [max_{c^{'}}h_\phi }(x_{i}^{u} )^{(j,c^{'})} > \tau] }{H\cdot W}
\end{equation}

The computation of the unsupervised loss term is determined by considering the pseudo-label and its corresponding quality. $L_{inter}$ denotes the unsupervised loss brought by inter-domain mixing, and $L_{intra}$ represents the unsupervised loss brought by intra-domain mixing.

\begin{equation}\label{eq:inter_domain_mixing_loss}
\scalebox{0.95}{$\displaystyle
L_{inter} = - \sum_{j=1}^{H\times W}\sum_{c=1}^{C}q_{i}(\hat{y}_{i}^{inter})^{(j,c)}\log g_{\theta}(x_{i}^{inter})^{(j,c)}
$}
\end{equation}

\begin{equation}\label{eq:intra_domain_mixing_loss}
\scalebox{0.95}{$\displaystyle
L_{intra} = - \sum_{j=1}^{H\times W}\sum_{c=1}^{C}q_{i}(\hat{y}_{i}^{intra})^{(j,c)}\log g_{\theta}(x_{i}^{intra})^{(j,c)}
$}
\end{equation}

In addition, we also perform supervised training on the source and labeled target data. $L_{s}$ denotes the supervised loss on source data, and $L_{t}$ represents the supervised loss on the labeled target data. The supervised loss term is computed as:

\begin{equation}\label{eq:supervised_loss_s}
L_{s} = - \sum_{j=1}^{H\times W}\sum_{c=1}^{C}(y_{i}^{s})^{(j,c)}\log g_{\theta}(x_{i}^{s})^{(j,c)}
\end{equation}
\begin{equation}\label{eq:supervised_loss_t}
L_{t} = - \sum_{j=1}^{H\times W}\sum_{c=1}^{C}(y_{i}^{t})^{(j,c)}\log g_{\theta}(x_{i}^{t})^{(j,c)}  
\end{equation}

Finally, the overall loss term is computed as:
\begin{equation}\label{eq:overall_loss}
    L_{final} = L_{s} + L_{t} + \lambda L_{inter} + \mu L_{intra}
\end{equation}
where $\lambda$ and $\mu$ denote the loss weight of inter and intra-domain mixing, respectively. 

\section{Experiments}
\subsection{Implementation Details}
\paragraph{Datasets.} For the target domain, we use the Cityscapes street scene dataset~\cite{cordts2016cityscapes} containing 2975 training and 500 validation images with resolution 2048×1024. All images are manually labeled by 19 semantic categories. For the source domain, we use either the GTA5 dataset~\cite{richter2016playing}, which contains 24,966 synthetic images with resolution 1914×1052, or the Synthia dataset~\cite{ros2016synthia}, which consists of 9,400 synthetic images with resolution 1280×760. As a common practice in SSDA, we randomly select different numbers of images, such as (100, 200, 500, 1000), from the whole target domain training set. These selected images are subsequently utilized as labeled target images, and the rest of target domain training set are utilized as unlabeled target images. In the setting of GTA5→Cityscapes, we consider the 19 common classes with Cityscapes dataset to train the models. In the setting of Synthia→Cityscapes, we train the models with the 16 common classes and report the 13-class mIoU on validation set. 

\noindent
\paragraph{Training.} In the default setting,
we employ the DeepLabv2~\cite{chen2017deeplab} with ResNet-101~\cite{he2016deep} pretrained on ImageNet-1k~\cite{deng2009imagenet}. We use the AdamW~\cite{loshchilov2017decoupled} as our optimizer with betas (0.9, 0.999) and weight decay 0.01. The learning rate is initially set to $6\times 10^{-5}$ for the encoder and $6\times 10^{-4}$ for decoder. The loss weights $\lambda$ and $\mu$ are set to 1.0 and 2.0, respectively. Similar to DAFormer~\cite{hoyer2022daformer}, learning rate warmup policy and rare class sampling are applied. During training, source images are rescaled to 760×1280 and target images to 512×1024, after which random crops of size 512×512 are extracted, and each mini-batch is composed of two source images, two labeled target images and two unlabeled target images, and we trained the network for total 40k iterations. Following DACS~\cite{tranheden2021dacs}, we adopt ClassMix~\cite{olsson2021classmix} as mixing technique and set $\alpha=0.99$ and $\tau=0.968$. Experiments were implemented in PyTorch~\cite{paszke2019pytorch} and trained on a single NVIDIA Tesla V100 GPU.

\subsection{Comparisons with State-of-the-Art Methods}
The performance is compared with existing state-of-the-art methods on UDA, SSL and SSDA settings. We employ mean intersection-over-union (mIoU) as the evaluation metric which is broadly adopted in semantic segmentation.

\noindent
\paragraph{GTA5 to Cityscapes.} The performance comparisons with several state-of-the-art methods on GTA5 to Cityscapes are shown in Table~\ref{GTA5_to_Cityscapes}. IIDM consistently achieves the best performance with varying number of labeled target images. 
Compared with the most recent SOTA UDA methods such as DIGA~\cite{shen2023diga} and CONFETI~\cite{li2023contrast}, IIDM can obtain above +6.8\% mIoU improvement with only 100 labeled target images.
This demonstrates the significance and efficacy of investigating SSDA, a research area that has long been overlooked. 
Additionally, The performance of purely supervised learning is not good due to its reliance on a limited amount of labeled target images. The scarcity of labeled target images impedes the model's capacity to learn generalized features on the target domain, and increases the risk of overfitting. 
Compared with SSL methods such as CowMix~\cite{french2019semi} and DST-CBC~\cite{feng2022dmt}, IIDM takes the overwhelming advantage. Compared with DusPerb, part of the latest SOTA SSL method UniMatch~\cite{yang2023revisiting}, IIDM still prevails. Both DusPerb and UniPerb are part of UniMatch. UniPerb proposes to inject perturbations on features, which is orthogonal to image-level perturbations and IIDM. Given that, we only compared with DusPerb.
Nonetheless, as the number of labeled target images escalates, the performance discrepancy between IIDM and DusPerb tends to attenuate.
With regard to the impact of more labeled target images, we have conducted a further analysis in ablation study.

Particularly, IIDM outperforms all other SSDA methods by a large margin in all settings. When provided with 100 labeled target images, the performance of IIDM is improved by +8.3\% mIoU compared with DLDM~\cite{chen2021semi1}, reaching 69.5\% mIoU, which also surpasses the performance of DLDM using 1000 labeled target images. Compared with ASS+SDSS~\cite{kim2022source}, which subsamples source domain data to generate a small-scale meaningful subset, IIDM achieves +13.4\% mIoU improvement through a more streamlined training procedure.

\begin{table}[htb]
  \centering
  \resizebox{0.99\columnwidth}{!}{
  \renewcommand{\arraystretch}{1.0}
  \begin{tabular}{l|c|ccccc}
    \hline
    \multirow{2}{*}{Type} & \multirow{2}{*}{Method} & \multicolumn{5}{c}{Labeled target images} \\
    & & 0 & 100 & 200 & 500 & 1000 \\
    \hline
    \multirow{3}{*}{UDA} & DACS & 52.1 & - & - & - & - \\
    & ProDA & 57.5 & - & - & - & - \\
    & DAFormer & 56.0 & - & - & - & - \\
    & CONFETI & 62.2 & - & - & - & - \\
    & DIGA & 62.7 & - & - & - & - \\
    \hline
    Sup. & Deeplabv2 & - & 41.9 & 47.7 & 55.5 & 58.6 \\
    \hline
    \multirow{3}{*}{SSL} & CowMix & - & 50.8 & 54.8 & 61.7 & 64.8 \\
    & DST-CBC & - & 48.7 & 54.1 & 60.6 & 63.2 \\
    & DusPerb* & - & 61.8 & 66.7 & 68.4 & 72.1 \\
    \hline
    \multirow{6}{*}{SSDA} & MME & - & 52.6 & 54.4 & 57.6 & 61.0 \\
    & ASS & - & 54.2 & 56.0 & 60.2 & 64.5 \\
    & ASS+SDSS & - & 56.1 & 58.4 & 62.8 & 65.7 \\
    & DLDM & - & 61.2 & 60.5 & 64.3 & 66.6 \\
    & ComplexMix & - & 60.1 & 62.9 & 65.7 & 66.8 \\
    & IIDM & - & \textbf{69.5} & \textbf{70.0} & \textbf{70.6} & \textbf{72.8} \\
    \hline
  \end{tabular}
  }
  \caption{Compared with the UDA, Supervised, SSL and SSDA methods on GTA5→Cityscapes. 
  }
  \label{GTA5_to_Cityscapes}
\end{table}

\begin{figure*}[t]
\centering
\includegraphics[width=0.8\textwidth]{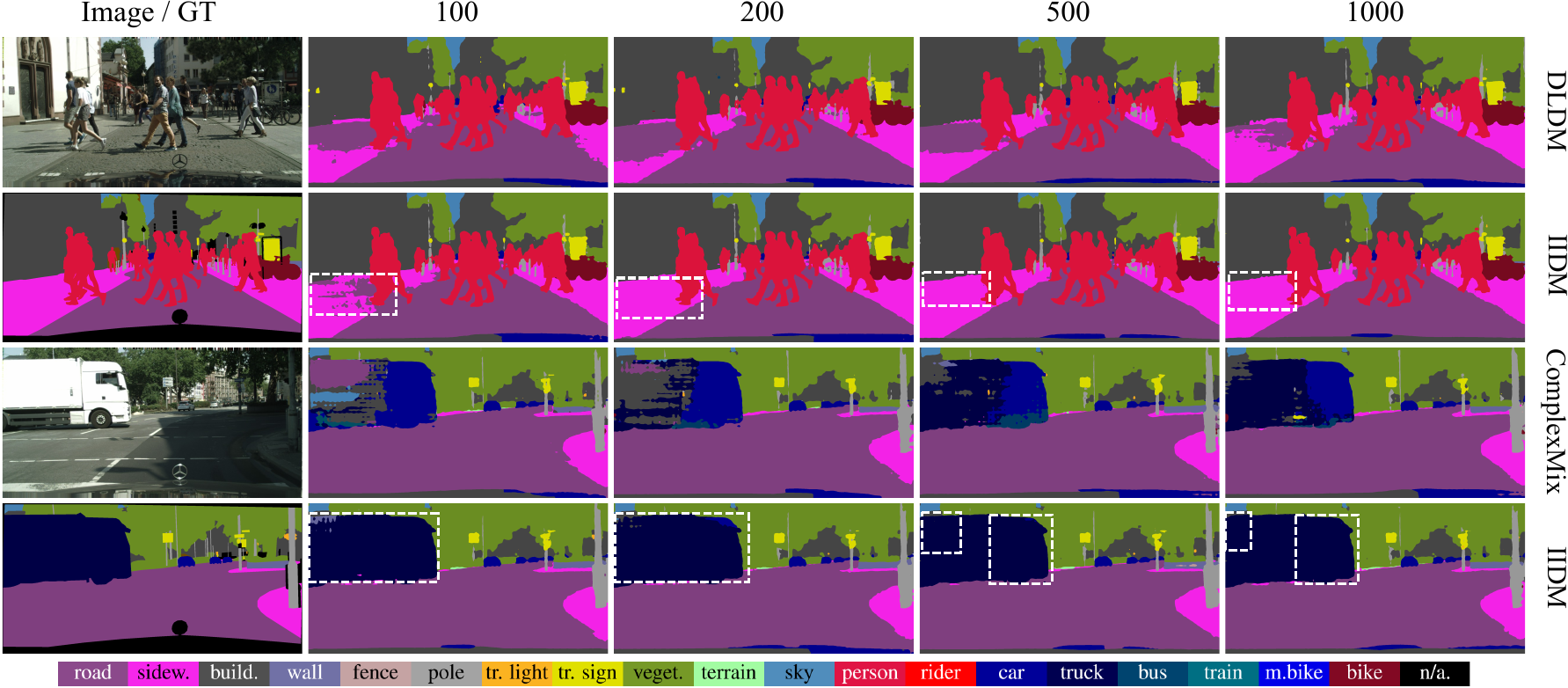}
\caption{Qualitative comparison of IIDM (row 2 and row 4) with previous methods (DLDM for row 1 and ComplexMix for row 3) on GTA5 to Cityscapes. IIDM better segments difficult classes such as sidewalk and truck.}
\label{visualization}
\end{figure*}

We visualize the segmentation results on 100, 200, 500 and 1000 labeled target images. As can be seen from Figure~\ref{visualization}, IIDM achieves smoother and more complete segmentation results than others using the same number of labeled target images. Moreover, IIDM yields precise results even when confronted with scarce labeled target images.

\noindent
\paragraph{Synthia to Cityscapes.} To further evaluating IIDM, we also compare the results with current state-of-the-art methods on the Synthia to Cityscapes setting, as shown in Table~\ref{Synthia_to_Cityscapes}. Following previous UDA and SSDA works~\cite{vu2019advent,chen2021semi1}, all the methods listed here are trained using 16 common classes between Synthia and Cityscapes and 13-class mIoU score is reported. It can be seen that IIDM still achieves the best performance on all ratios of labeled target images. When the number of labeled target images is 100, the performance of IIDM is improved by +3.6\% mIoU compared with ComplexMix, reaching 74.2\% mIOU, and significantly surpassing the performance of supervised learning using 1000 labeled target images. The consistent performance improvements across different datasets clearly demonstrates the effectiveness of our proposed method.

\begin{table}[htb]
  \centering
  \resizebox{0.99\columnwidth}{!}{
  \renewcommand{\arraystretch}{1.0}
  \begin{tabular}{l|c|ccccc}
    \hline
    \multirow{2}{*}{Type} & \multirow{2}{*}{Method} & \multicolumn{5}{c}{Labeled target images} \\
    & & 0 & 100 & 200 & 500 & 1000 \\
    \hline
    \multirow{3}{*}{UDA} & DACS & 54.8 & - & - & - & - \\
    & ProDA & 62.0 & - & - & - & - \\
    & DAFormer & 58.2 & - & - & - & - \\
    & DiGA & 67.9 & - & - & - & - \\
    \hline
    Sup. & Deeplabv2 & - & 53.0 & 58.9 & 61.0 & 67.5 \\
    \hline
    \multirow{3}{*}{SSL} & CowMix & - & 61.3 & 66.7 & 71.1 & 73.0 \\
    & DST-CBC & - & 59.7 & 64.3 & 68.9 & 70.5 \\
    & DusPerb* & - & 68.4 & 71.4 & 74.2 & 76.1 \\
    \hline
    \multirow{6}{*}{SSDA} & MME & - & 59.6 & 63.2 & 66.7 & 68.9 \\
    & ASS & - & 62.1 & 64.8 & 69.8 & 73.0 \\
    & ASS+SDSS & - & 64.2 & 66.4 & 69.9 & 73.5 \\
    & DLDM & - & 68.4 & 69.8 & 71.7 & 74.2 \\
    & ComplexMix & - & 70.6 & 71.8 & 72.6 & 74.0 \\
    & IIDM & - & \textbf{74.2} & \textbf{76.4} & \textbf{77.0} & \textbf{78.8} \\
    \hline
  \end{tabular}
  }
  \caption{Comparisons on Synthia→Cityscapes. Here we train the models with the 16 common classes and report the 13-class mIoU.
  }
  \label{Synthia_to_Cityscapes}
\end{table}
\vspace{-10pt}

\subsection{Results with SOTA UDA methods}
The state-of-the-art unsupervised domain adaptation (UDA) approach, which incorporates MIC with HRDA, attains 75.9\% mIoU on GTA5 to Cityscapes setting. In pursuit of enhanced SSDA results, we incorporate the advanced training techniques employed in HRDA, such as multi-resolution self-training, increased training resolution and more robust backbone architectures. Furthermore, we conduct equitable comparisons with state-of-the-art UDA methods, such as DIGA+HRDA and MIC+HRDA, to thoroughly assess the effectiveness of our proposed method. As shown in Table~\ref{GTA5_to_Cityscapes_SOTA}, IIDM can achieve better performance with HRDA, attains 77.8\% mIoU with only 100 labeled target images. Furthermore, IIDM is orthogonal to MIC, thereby offering additional performance improvements (+0.8\% mIoU).

\begin{table}[htb]
  \centering
  \resizebox{0.99\columnwidth}{!}{
  \renewcommand{\arraystretch}{1.0}
  \begin{tabular}{l|c|ccccc}
    \hline
    \multirow{2}{*}{Type} & \multirow{2}{*}{Method} & \multicolumn{5}{c}{Labeled target images} \\
    & & 0 & 100 & 200 & 500 & 1000 \\
    \hline
    \multirow{3}{*}{UDA} 
    & HRDA  & 73.8 & - & - & - & - \\
    & DIGA + HRDA  & 74.3 & - & - & - & - \\
    & MIC + HRDA  & 75.9 & - & - & - & - \\
    \hline
    \multirow{2}{*}{SSDA}
    & IIDM + HRDA   & - & 77.8 & 78.9 & 79.6 & 81.1 \\
    & IIDM + MIC + HRDA& - & \textbf{78.6} & \textbf{79.5} & \textbf{80.2} & \textbf{81.2} \\
    \hline
  \end{tabular}
  }
  \caption{Results with SOTA UDA methods by leveraging the advanced training techniques employed in HRDA.}
  \label{GTA5_to_Cityscapes_SOTA}
\end{table}
\vspace{-10pt}

\subsection{Ablation Study}
To further demonstrate the effectiveness of IIDM, we conduct ablation studies on the GTA5 to Cityscapes setting.

\paragraph{Component Ablation.} In the proposed IIDM, we incorporate four distinct loss functions, including a supervised loss $L_{s}$ applied to the source data, a supervised loss $L_{t}$ applied to the labeled target data, an unsupervised loss $L_{inter}$ related to inter-domain mixing, and an unsupervised loss $L_{intra}$ associated with intra-domain mixing. Different losses represent different components, and the performance of different combinations is shown in Table~\ref{Ablation1}. As can be observed, the utilization of solely supervised loss (row 1, row 2 and row 3) yields poor performance particularly when labeled target data is scarce, 
and the learned feature center may exhibit a significant disparity with the target domain.
The combination of $L_{t}$ and $L_{intra}$ (row 4) represents the SSL method. However, in the absence of source domain data to serve as a basis, the performance is compromised (59.1\% vs 69.5\%), particularly when the number of labeled target images is limited. The ablation experiments of inter and intra-domain mixing are denoted by row 5, row 6 and row 7. The performance achieved by exclusively employing intra-domain mixing (row 6) surpasses that of using only inter-domain mixing (row 5), indicating the importance of intra-domain mixing in enriching the target domain space.This is because the center of intra-domain mixed data is closer to the center of the target domain than that of the inter-domain mixed data, as described in Figure~\ref{motivation}. Finally, incorporating four distinct losses achieves the best results and outperforms others (row 5 and row 6) by a large amount. Learning with both inter and intra-domain mixing enables the extraction of features that are domain-invariant in a more abundant enhanced target domain space.

\begin{table}[htb]
\centering
\resizebox{0.99\columnwidth}{!}{
\renewcommand{\arraystretch}{1.0}
\begin{tabular}{ccccccccc}
\hline
$L_{s}$ & $L_{t}$ & $L_{inter}$ & $L_{intra}$ & 0 & 100 & 200 & 500 & 1000  \\
\hline
\checkmark &  &  &  & 37.3 & - & - & - & -  \\
 & \checkmark &  &  & - & 48.7 & 52.3 & 60.1 & 67.3  \\
\checkmark & \checkmark &  &  & - & 48.5 & 53.5 & 60.9 & 66.7  \\
 & \checkmark &  & \checkmark & - & 59.1 & 65.5 & 68.2 & 70.5  \\
\checkmark & \checkmark & \checkmark &  & - & 65.7 & 67.5 & 67.8 & 70.5  \\
\checkmark & \checkmark &  & \checkmark & - & 66.9 & 68.4 & 70.0 & 71.7  \\
\checkmark & \checkmark & \checkmark & \checkmark & - & \textbf{69.5} & \textbf{70.0} & \textbf{70.6} & \textbf{72.8} \\
\hline
\end{tabular}
}
\caption{Ablation study on the effectiveness of various losses. Here different losses represent different components.
}
\label{Ablation1}
\end{table}
\vspace{-10pt}

\paragraph{Intra and inter-domain mixing strategies.} In the proposed IIDM, we employ a straightforward domain mixing strategy, wherein the inter and intra-domain mixing are applied on the same unlabeled target image and learns via two separate streams (\textbf{one $x^{u}$ two streams}). However, alternative options exist, such as 1) applying on two distinct unlabeled target images and learning via two separate streams (\textbf{two $x^{u}$ two streams}) or 2) applying on the same unlabeled target image and learning via a single stream (\textbf{one $x^{u}$ one stream}). We conduct ablation experiments on the domain mixing strategies, as presented in Table~\ref{Ablation2}. It is observed that “one $x^{u}$ two streams” performs best. 
While all three strategies incorporate both inter and intra-domain mixing, applying them on the same image enables the network to aware the two domain gaps simultaneously, which benefits the network to learn more consistent feature representation across domains.

\begin{table}[htb]
\centering
\resizebox{0.82\columnwidth}{!}{
\renewcommand{\arraystretch}{1.0}
\begin{tabular}{c|cccc}
\hline
Method & 100 & 200 & 500 & 1000  \\
\hline
two $x^{u}$ two streams & 66.8 & 67.4 & 69.7 & 72.0  \\
one $x^{u}$ one stream & 67.2 & 68.2 & 69.0 & 71.6  \\
one $x^{u}$ two streams & \textbf{69.5} & \textbf{70.0} & \textbf{70.6} & \textbf{72.8}    \\
\hline
\end{tabular}
}
\caption{Ablation study on various domain mixing strategies.}
\label{Ablation2}
\end{table}
\vspace{-10pt}


\paragraph{Loss weights $\lambda$ and $\mu$.} In the proposed IIDM, loss weights $\lambda$ and $\mu$ represent the loss ratio of inter and intra-domain mixing respectively. We ablate this two hyper-parameters in Table~\ref{Ablation3}. The model achieves the optimal performance when $\lambda$ and $\mu$ are set to 1 and 2, respectively.
When either $\lambda$ or $\mu$ is set to 0.1 (row 1 and row 5), a noticeable decrease in performance is observed, and the most significant drop occurs when the number of labeled target images is 100. 
Through a comparison of the results presented in row 2, row 3, and row 4, it becomes evident that greater emphasis should be placed on the intra-domain mixing, which has been totally disregarded in previous semi-supervised domain adaptation (SSDA) methods. When labeled target images is sufficient, performance is more robust to the two hyper-parameters.

\begin{table}[ht]
\centering
\resizebox{0.64\columnwidth}{!}{
\renewcommand{\arraystretch}{1.0}
\begin{tabular}{cc|cccc}
\hline
$\lambda$ & $\mu$ & 100 & 200 & 500 & 1000  \\
\hline
0.1 & 1 & 66.1 & 67.9 & 69.8 & 72.1  \\
2 & 1   & 68.5 & 69.5 & 70.5 & 71.7  \\
1 & 1   & 68.7 & 69.2 & 70.4 & 72.5  \\
1 & 2   & \textbf{69.5} & \textbf{70.0} & \textbf{70.6} & \textbf{72.8}  \\
1 & 0.1 & 67.3 & 68.1 & 68.7 & 71.4  \\
\hline
\end{tabular}
}
\caption{Ablation study on the efficacy of weights $\lambda$ and $\mu$.}
\label{Ablation3}
\end{table}
\vspace{-10pt}

\paragraph{More labeled target images.} We report the experimental results by gradually increasing the number of labeled target images over 1000 in Table~\ref{GTA5_to_Cityscapes_2475}. As the number increases, the performance improvement of IIDM diminishes, while the performance of fully supervised and DusPerb exhibits a more pronounced improvement. Specifically, when the number of labeled target images reaches 2475, DusPerb surpasses IIDM, indicating a diminishing effect of domain adaptation. 
This is because the domain gap caused by random sampling decreases, and the data center of the labeled target gradually aligns with the data center of the entire target domain. Consequently, utilizing source domain data when the labeled target images reach a sufficient size may become redundant. However, the source data is valuable in scenarios where labeled target data is either unavailable or limited in quantity.
Since there is no unlabeled target data in column 2975, DusPerb and IIDM can not be implemented in that case.



\begin{table}[htb]
  \centering
  \resizebox{0.9\columnwidth}{!}{
  \renewcommand{\arraystretch}{1.0}
  \begin{tabular}{l|c|ccccc}
    \hline
    \multirow{2}{*}{Type} & \multirow{2}{*}{Method} & \multicolumn{5}{c}{Labeled target images} \\
    & & 1000 & 1488 & 1975 & 2475 & 2975 \\
    \hline
    Sup. & Deeplabv2 & 58.6 & 69.4 & 71.1 & 71.9 & 72.6 \\
    \multirow{1}{*}{SSL}
    & DusPerb* & 72.1 & 72.8 & 73.2 & \textbf{74.2} & -  \\
    \multirow{1}{*}{SSDA}
    & IIDM & \textbf{72.8} & \textbf{73.0} & \textbf{73.3} & 73.7 & - \\
    \hline
  \end{tabular}
  }
  \caption{Additional experiments using more labeled data.}
  \label{GTA5_to_Cityscapes_2475}
\end{table}
\vspace{-10pt}

\section{Conclusion}
In this paper, we reveal the importance of properly utilizing intra-domain information in SSDA for semantic segmentation, which is always overlooked by previous SSDA methods. Based on our observation, an SSDA framework that utilizing both inter and intra-domain mixing (IIDM) is proposed. Learning with inter and intra-domain mixing enables the network to better aware the domain gap and learn more consistent feature representations across domains. Different domain mixing strategies are also explored for better enriching the information of target domain. Extensive experiments demonstrates the effectiveness of the proposed IIDM which outperforms previous SSDA methods with a large margin. For instance, IIDM respectively improves the state-of-the-art SSDA performance by +8.3\% and +3.6\% mIoU on GTA5→Cityscapes and Synthia→Cityscapes. We hope that, due to its simplicity, IIDM can be used as part of future SSDA methods, and garnering more attention from researchers in this practical setting of SSDA.

\clearpage
\bibliography{aaai24}

\end{document}